\documentclass[letterpaper]{article} 
\usepackage{aaai2026}  
\usepackage{times}  
\usepackage{helvet}  
\usepackage{courier}  
\usepackage[hyphens]{url}  
\usepackage{graphicx} 
\usepackage{natbib}  
\usepackage{caption} 
\usepackage{algorithm}
\usepackage{xcolor}
\usepackage{algpseudocode}
\usepackage{amsmath}
\usepackage{amssymb}
\usepackage{bibentry}
\usepackage[inline]{enumitem}
\usepackage{multirow}
\usepackage{booktabs}
\usepackage{array}

\usepackage{arydshln}
\usepackage{newfloat}
\usepackage{listings}

\DeclareCaptionStyle{ruled}{labelfont=normalfont,labelsep=colon,strut=off} 
\floatstyle{ruled}
\newfloat{listing}{tb}{lst}{}
\floatname{listing}{Listing}
\newcommand{\heading}[1]{\vspace*{0.8mm}\noindent\textbf{#1.}}

\title{Thinking Forward and Backward:\\ Multi-Objective Reinforcement Learning for Retrieval-Augmented Reasoning}

\author {
    Wenda Wei\textsuperscript{\rm 1,\rm 2},
    Yu-An Liu\textsuperscript{\rm 1,\rm 2},
    Ruqing Zhang\textsuperscript{\rm 1,\rm 2}\thanks{Jiafeng Guo and Ruqing Zhang are the corresponding authors.},
    Jiafeng Guo\textsuperscript{\rm 1,\rm 2}\footnotemark[1],
    Lixin Su\textsuperscript{\rm 3},
    Shuaiqiang Wang\textsuperscript{\rm 3},
    Dawei Yin\textsuperscript{\rm 3},
    Maarten de Rijke\textsuperscript{\rm 4},
    Xueqi Cheng\textsuperscript{\rm 1,\rm 2}    
}

\affiliations {
    \textsuperscript{\rm 1}State Key Laboratory of AI Safety, Institute of Computing Technology, Chinese Academy of Sciences, Beijing, China\\
    \textsuperscript{\rm 2}University of Chinese Academy of Sciences, Beijing, China\\
    \textsuperscript{\rm 3}Baidu Inc., Beijing, China\\
    \textsuperscript{\rm 4}University of Amsterdam, Amsterdam, The Netherlands\\
    \{weiwenda25z, liuyuan21b, zhangruqing, guojiafeng, cxq\}@ict.ac.cn, \{sulixin, wangshuaiqiang, yindawei02\}@baidu.com, m.derijke@uva.nl 
}

\begin{document}

\maketitle

\begin{abstract}
Retrieval-augmented generation (RAG) has proven to be effective in mitigating hallucinations in large language models, yet its effectiveness remains limited in complex, multi-step reasoning scenarios.
Recent efforts have incorporated search-based interactions into RAG, enabling iterative reasoning with real-time retrieval. 
Most approaches rely on outcome-based supervision, offering no explicit guidance for intermediate steps. 
This often leads to reward hacking and degraded response quality.  
We propose Bi-RAR, a novel retrieval-augmented reasoning framework that evaluates each intermediate step jointly in both forward and backward directions.
To assess the information completeness of each step, we introduce a bidirectional information distance grounded in Kolmogorov complexity, approximated via language model generation probabilities. 
This quantification measures both how far the current reasoning is from the answer and how well it addresses the question. 
To optimize reasoning under these bidirectional signals, we adopt a multi-objective reinforcement learning framework with a cascading reward structure that emphasizes early trajectory alignment.
Empirical results on seven question answering benchmarks demonstrate that Bi-RAR surpasses previous methods and enables efficient interaction and reasoning with the search engine during training and inference. \looseness=-1
\end{abstract}

\vspace{-3mm}
\section{Introduction}

Retrieval-augmented generation (RAG) \cite{lewis2020retrieval}  has emerged as a prominent framework for mitigating hallucination in large language models (LLMs) \cite{achiam2023gpt,team2024gemini,LLMSurvey}.

\heading{Integrating RAG with reasoning} While basic RAG methods are effective, they often struggle in real-world scenarios involving complex and heterogeneous data \cite{gao2023retrieval} that require multi-hop retrieval \cite{hendrycks2020measuring}. 
To address these limitations, recent research has increasingly focused on enhancing RAG with advanced reasoning capabilities. 
Specifically, LLMs can be prompted or trained to incorporate external tools, such as search engines, into a more dynamic and iterative reasoning process \cite{zhao2024retrieval}.

A representative paradigm is Search-R1 \cite{jin2025searchr1}, which has achieved strong performance across a range of question answering benchmarks.
The key idea is to optimize LLM reasoning trajectories through multi-turn search interactions, using retrieved token masking to enable reinforcement learning (RL) training.
The success of Search-R1 is largely attributed to its outcome-based reward function based on the correctness of the final answer. 
However, this form of supervision lacks explicit feedback for intermediate reasoning steps, making it difficult to control the reasoning process throughout. 
As a result, such optimizations may induce in-context reward hacking, where the model generates unnecessarily long or inefficient reasoning chains. These extended chains can accumulate hallucinations and ultimately compromise the final response. 
\emph{Can we precisely supervise the information understanding at each reasoning step?}

\heading{Beyond unidirectional reasoning} 
Cognitive research has shown that humans reason not only in a forward deduction, from problem to solution, which reflects how the brain plans over unknown information, but also in a backward deduction, from solution to problem \cite{hawes2012experience}.
Bidirectional deductive reasoning enables the brain to evaluate the reliability of known information and to plan toward the unknown information, ensuring a reasoning process that bridges the gap between the question and the answer.
A recent study has also demonstrated that LLMs can similarly benefit from integrating forward and backward reasoning in complex tasks \cite{chen2024reverse}.
Inspired by these findings, \emph{we explore optimizing each step through top-down planning over unknown information and bottom-up evaluation of known information.}

\heading{Our method: RAG with bidirectional reasoning} 
We propose a novel retrieval-augmented reasoning framework, Bi-RAR, which dynamically evaluates each reasoning step through both forward and backward guidance to determine whether it provides sufficient support for task-solving.
To achieve this, we need to address two key challenges.

First, \emph{how to quantify the information completeness of each step from both forward and backward perspectives?} 
Kolmogorov complexity \cite{li1993introduction}, a foundational concept in information theory, defines the amount of information required to describe an object.
Building on this, information distance \cite{bennett1998information,vitanyi2009normalized,zhang2007information} provides a universal, domain-agnostic metric for measuring the similarity between objects and has successfully been applied across a variety of domains \cite{li1993introduction,zhang2007information,li2008answer}. 
In this work, we adopt a conditional normalized information distance under specified condition patterns.
For forward completeness, we measure how far the current reasoning context is from the final answer; for backward completeness, we assess how well it addresses the input question.
By approximating Kolmogorov complexity via language model generation probabilities, we estimate the information distance in both directions, thereby capturing the information completeness of each step.

Second, \emph{how to optimize step-wise reasoning using forward-backward information distances?} 
Given the effectiveness of RL \cite{kaelbling1996reinforcement} in sequential decision-making, and the bidirectional signals introduced above, we propose to use multi-objective RL methods \cite{roijers2013survey,li2020deep} to explore the entire preference space.
Concretely, we first design a cascading reward structure that prioritizes the early establishment of correct reasoning directions, based on the forward and backward information distances, respectively.
These two reward signals serve as the primary supervision for guiding RL optimization. 
We then train specialized models with their respective rewards independently, using group relative policy optimization (GRPO) \cite{shao2024deepseekmath}. 
During training, the model progressively learns to perform accurate and efficient multi-step reasoning, dynamically determining whether and how to invoke the search engine at each step, in order to optimize the forward or backward objective.
Finally, we obtain a balanced solution through weight-space interpolation, which enables task-specific optimization by selecting appropriate interpolation settings.

Experiments conducted on seven widely-used question answering benchmarks demonstrate that Bi-RAR achieves strong overall performance, with particularly notable improvements of 18.2\% (Qwen2.5-3B-Instruct) and 8.3\% (Qwen2.5-3B-Base) over the strongest baseline Search-R1 \cite{jin2025searchr1}, while using only one-fourth of Search-R1’s training data.
Further analyses show that Bi-RAR is more effective in both training and inference.

\vspace{-2mm}
\section{Preliminaries}
In this section, we review Search-R1 \cite{jin2025searchr1}, a representative method for enhancing retrieval-augmented generation with reasoning capabilities.

\heading{Search-R1} 
Recent advances, such as Search-R1, extend RAG to support multi-step reasoning with interleaved retrieval. 
In this paradigm, given a question $Q$, the LLM generates a reasoning trajectory $T = \{T_1, T_2, \ldots, T_n\}$. At each step $i$, the LLM
\begin{enumerate*}[label=(\roman*)] 
\item first generates a reasoning step $T_i$ based on the current information;
\item then, it issues a search query and retrieves relevant documents; and
\item finally, it judges whether to move on to the next reasoning step $T_{i+1}$ or generate the final answer $A$ based on the current content.
\end{enumerate*}  
The process alternates between reasoning and search.

During training, RL is employed to encourage the LLM to interact effectively with the search engine. 
A reward function $r_\phi$ evaluates the correctness of the final answer extracted from the model’s output. 
To ensure that the LLM generates valid and stable search engine calls, a structured prompting template is adopted to structure the
model's output into three parts in an iterative fashion: reasoning process, search engine calling function, and the answer. 
Specifically, the RL policy $\pi_\theta$ is optimized by:
\begin{equation}
\begin{aligned}
\max_{\pi_\theta} {} & \mathbb{E}_{x \sim \mathcal{D}, y \sim \pi_\theta(\cdot|x;\mathcal{R})} 
\left(r_\phi - {}\right.\\
& 
\mbox{}\hspace*{15mm}
\left.
\beta D_{\text{KL}}[\pi_\theta(y|x;\mathcal{R}) \| \pi_{\text{ref}}(y|x;\mathcal{R})]\right),
\end{aligned}
\end{equation}
where $\pi_{\text{ref}}$ is a reference policy, $D_{\text{KL}}$ is the
KL-divergence measure, $\beta$ controls the strength of the KL penalty, $\mathcal{R}$ is the search engine, $x$ are input samples from dataset $\mathcal{D}$, and $y$ are the generated outputs. 
And $\pi_\theta(\cdot|x;\mathcal{R})$ denotes the policy that generates text interleaved with the search engine. 


\heading{Discussion}
Search-R1 aims to teach LLMs when and how to interact with a search engine during reasoning.
However, its outcome-based supervision focuses solely on the correctness of the final answer, which can easily lead to reward hacking by the model.
This behavior is characterized by the LLM issuing a large number of loosely relevant queries in an attempt to improve the answer through excessive retrieval, rather than through deliberate and coherent reasoning. 
As shown in Section~\ref{sec:Efficiency Analysis}, this not only reduces efficiency due to unnecessarily lengthy reasoning trajectories, but also introduces redundant information that may accumulate across steps, increasing the risk of hallucinated content and ultimately derailing the reasoning process.
In this paper, we explore how to provide fine-grained guidance at each intermediate step of the reasoning trajectory to support more efficient and accurate retrieval-augmented reasoning approach.

\begin{figure}[t]
    \centering
    \includegraphics[width=\linewidth]{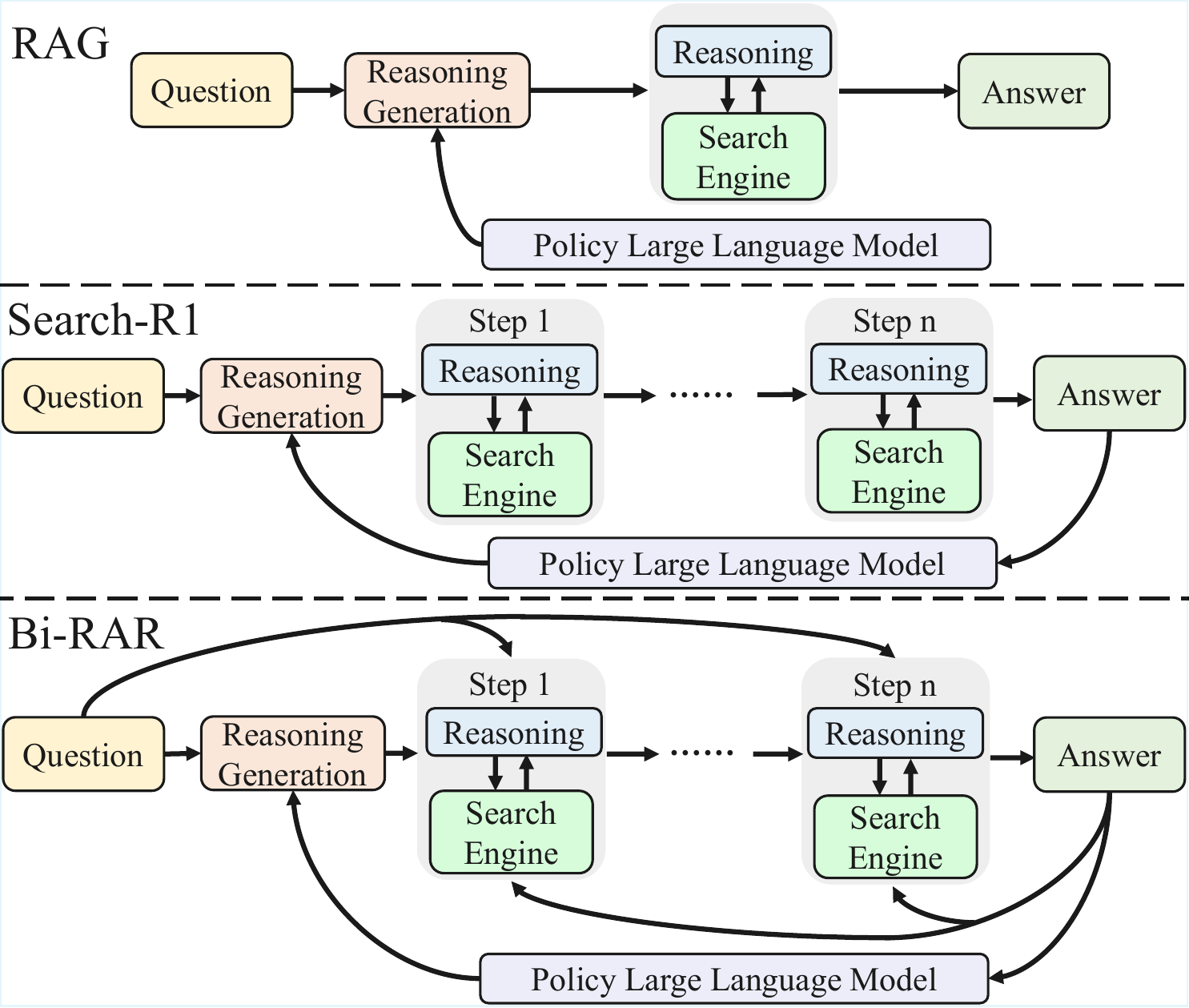}
    \caption{Framework of Bi-RAR compared with typical  RAG \cite{lewis2020retrieval} and Search-R1 \cite{jin2025searchr1}.}
    \label{fig: framework}
    \vspace{-2mm}
\end{figure}

\section{Method}
\subsection{Overview}
In this section, we present Bi-RAR, a retrieval-augmented reasoning framework that uses bidirectional reasoning to optimize the intermediate steps in answering complex questions.
As illustrated in Figure~\ref{fig: framework}, our approach comprises two main components:
\begin{enumerate*}[label=(\roman*)]
\item Bidirectional information quantification: at each reasoning step, we evaluate the information distance to both the final answer and the original question, assessing step-wise information completeness; and 
\item Multi-objective optimization: these distances serve as bidirectional rewards, we use a multi-objective strategy to balance answer-seeking and question-grounding, guiding the model toward well-structured reasoning.
\end{enumerate*}

\vspace{-1mm}
\subsection{Bidirectional information quantification}
\heading{Motivation}
Effective multi-step reasoning requires fine-grained supervision signals that can evaluate the quality of each intermediate step.
The central challenge is to quantify the information completeness of each step, i.e., assessing whether a step meaningfully advances problem-solving while remaining faithful to the original question.

To tackle this, we draw inspiration from Kolmogorov complexity theory \cite{li1993introduction}, which offers a domain-independent, information-theoretic foundation for assessing semantic relevance based on minimal description length. 
We propose a mechanism to provide efficient feedback during LLM training by quantifying the information distance between each reasoning step and both the final answer and the original question.

\heading{Information distance based on Kolmogorov complexity}
Kolmogorov complexity \cite{li1993introduction} measures the amount of information contained in an individual object.
Given a string $a$, its Kolmogorov complexity $K(a)$ is defined as the length of the shortest binary program that outputs $a$ under a fixed universal computational model.
The conditional complexity $K(a|c)$ refers to the shortest program that generates $a$ given some auxiliary input string $c$, capturing the information in $a$ that is not already present in $c$.
More generally, $K(a|b,c)$ quantifies the information required to produce $a$ when both strings $b$ and $c$ are known. 

Here, we adopt the normalized information distance (NID) \cite{zhang2007information}, which uses Kolmogorov complexity to define a universal, context-aware similarity metric between two pieces of content. 
Formally, given two strings $a$ and $b$ with background context string $c$, the conditional normalized information distance between $a$ and $b$ is: 
\begin{equation}
\label{dis}
d(a, b|c) = \frac{\min\{K(a|b,c), K(b|a,c)\}}{\min\{K(a|c), K(b|c)\}}  .
\end{equation}
Since Kolmogorov complexity is uncomputable \cite{li1993introduction}, we approximate it in Eq.~(\ref{dis}), using the generation probabilities of a language model: 
\begin{equation}
\begin{aligned}
K(u|v) &\approx -\log_2P_{\text{LM}}(u|v), \\
K(u|v,w) &\approx -\log_2P_{\text{LM}}(u|v,w),
\end{aligned}
\label{kol}
\end{equation}
where $P_{\text{LM}}(u|v)$ and $P_{\text{LM}}(u|v,w)$ denote the likelihood of generating $u$ given the context $v$ or the joint context $(v,w)$, as computed by a language model.
This approximation is grounded in Shannon's information theory \cite{shannon1948mathematical}, aligning with the concept of entropy. 
In our implementation, we use Qwen2.5-3B \cite{qwen2025qwen25technicalreport} as the underlying language model. 
By employing the same language model as the generative model, this approximation can reasonably estimate the Kolmogorov complexity, which is the shortest program length required to generate the object given the contextual information.

\begin{figure}[t]
    \centering
    \includegraphics[width=\linewidth]{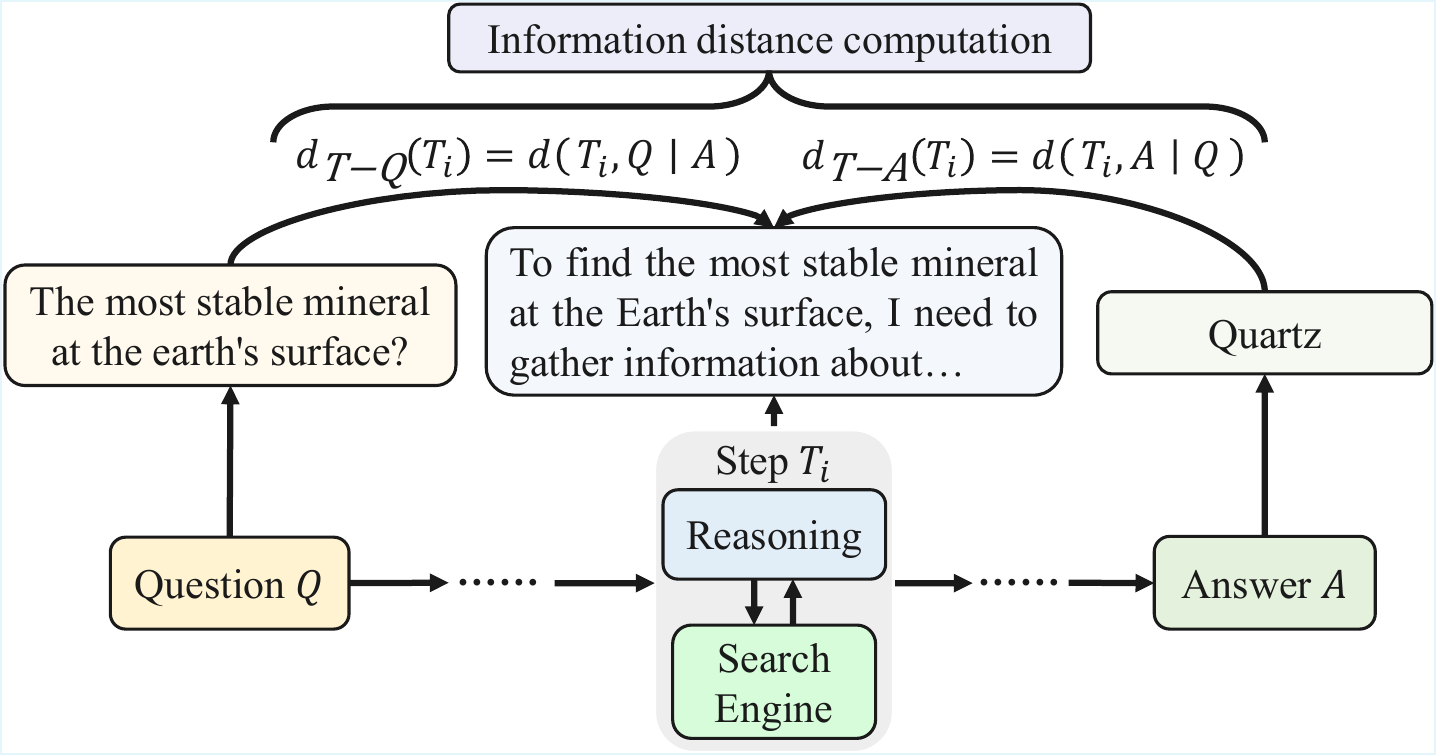}
    \caption{Sample of bidirectional distances computation.}
    \vspace{-2mm}
    \label{fig: bi-dis}
 
\end{figure}

\heading{Bidirectional distances}
Building on the normalized information distance defined in Eq.~(\ref{dis}), with conditional Kolmogorov complexity approximated by Eq.~(\ref{kol}), we propose two complementary metrics to quantify the bidirectional informativeness of each reasoning step, which is generated by the LLM based on the question, previous reasoning steps,  and retrieved documents.
As shown in Figure~\ref{fig: bi-dis}, for each step $T_i$, we compute: 
\begin{equation}
\label{dual_dis}
\mbox{}
\hspace*{-2mm}
\begin{cases}
d_{\text{T-A}}(T_i) = d(T_i, A \mid Q), & \text{step-to-answer distance} \\
d_{\text{T-Q}}(T_i) = d(T_i, Q \mid A), & \text{step-to-question distance}.
\end{cases}
\end{equation} 
For each reasoning step $T_i$, these two distances reflect different aspects of information completeness:
\begin{enumerate}
\item \emph{Step-to-answer distance} $d_{\text{T-A}}(T_i)$ quantifies how much the current step $T_i$ contributes toward the final answer $A$, indicating its solution progress; and 
\item \emph{Step-to-question distance} $d_{\text{T-Q}}(T_i)$ assesses how well $T_i$ remains grounded in the original question $Q$, ensuring contextual relevance and fidelity to the task.
\end{enumerate}
This bidirectional formulation enables a comprehensive evaluation of each reasoning step, allowing the model to dynamically balance between deep exploration and consistent alignment with the question.

\subsection{Multi-objective optimization with RL} \label{Sec: multi-sec}
\heading{Motivation}
In this work, the LLM performs multi-step reasoning guided by bidirectional distances at each step. This frames the task as a multi-objective, multi-step sequential decision problem.
To optimize this, we adopt RL to train the entire inference sequence, with three main components:
\begin{enumerate*}[label=(\roman*)]
\item Designing bidirectional rewards derived from the information distances to supervise training;
\item Independently training models with the search engine, each optimized for a single reward, to mitigate conflicts between forward and backward objectives; and
\item Combining the two models via weighted interpolation to obtain a balanced solution that guides the model to generate reasoning steps both relevant to the question and progressively closer to the correct answer. 
\end{enumerate*}

\heading{Bidirectional rewards design}
Based on the computed bidirectional distances, we define corresponding bidirectional reward functions.
To account for the varying importance of reasoning steps, we introduce a cascading reward structure that prioritizes early establishment of correct reasoning directions. 
Specifically, the forward reward $R_{\text{forward}}$ and backward reward $R_{\text{backward}}$ are defined as:
\begin{equation}
\label{rewardans}
R_{\text{forward}} = \mathbb{1}[\text{correct}] \cdot \sum_{i=1}^{n} \left[ \prod_{j=1}^{i-1} (1 - r_j^{\text{T-A}}) \right] r_i^{\text{T-A}},
\end{equation}
\begin{equation}
\label{rewardque}
R_{\text{backward}} = \mathbb{1}[\text{correct}] \cdot \sum_{i=1}^{n} \left[ \prod_{j=1}^{i-1} (1 - r_j^{\text{T-Q}}) \right] r_i^{\text{T-Q}} ,
\end{equation}
where $\mathbb{1}[\text{correct}]$ equals 1 if the final answer is correct, and 0 otherwise;
and 
\begin{equation}
r_i^{\text{T-A}} = e^{-d_{\text{T-A}}(T_i)}, \quad r_i^{\text{T-Q}} = e^{-d_{\text{T-Q}}(T_i)},
\end{equation}
represent the rewards derived from the step-to-answer and step-to-question distances at step $i$.

The exponential mapping ensures rewards increase as distances decrease, normalizing values between 0 and 1.
The cascading factor $\prod_{j=1}^{i-1}(1 - r_j)$ diminishes the contribution of later steps if earlier steps already show strong alignment, thereby encouraging efficient reasoning paths that establish correct directions as early as possible.

\heading{Independent training with the search engine}
To mitigate convergence issues from conflicting optimization objectives between the two rewards in early training, we initialize and train two models independently from the same pretrained checkpoint. Specifically,
\begin{enumerate*}[label=(\roman*)]
\item $\theta_{\text{forward}}$ is only optimized for the forward reward $R_{\text{forward}}$;
\item $\theta_{\text{backward}}$ is only optimized for the backward reward $R_{\text{backward}}$.
\end{enumerate*}
During training, the model can autonomously interact with the retriever at each reasoning step based on its current needs.
RL guides the model to perform accurate  multi-step reasoning and streamlined retrieval to optimize the forward or backward objective.

Each model is trained using group relative policy optimization (GRPO) \cite{shao2024deepseekmath}, which enhances training stability by employing group-wise baselines instead of value networks. For each input question $x$, we sample $G$ candidate responses $\{y_i\}_{i=1}^G$ from the current policy $\pi_{\text{old}}$ with the search engine $\mathcal{R}$, and optimize the objective: 
\begin{equation}
\begin{aligned}
&J_{\text{GRPO}}(\theta) = \mathbb{E}_{x \sim \mathcal{D}, \{y_i\}_{i=1}^G \sim \pi_{\text{old}}(\cdot|x;\mathcal{R})} \big[ \frac{1}{G} \sum_{i=1}^{G} \frac{1}{|y_i|} \sum_{t=1}^{|y_i|} \\
& \min\left( r_{i,t}(\theta) \hat{A}_{i,t}, \right. \left. \text{clip}\left( r_{i,t}(\theta), 1-\epsilon, 1+\epsilon \right) \hat{A}_{i,t} \right)\\
&- \beta D_{\text{KL}}[\pi_\theta || \pi_{\text{ref}}] \big],
\end{aligned}
\end{equation}
\begin{equation}
r_{i,t}(\theta) = \frac{\pi_\theta(y_{i,t} | x, y_{i,<t};\mathcal{R})}{\pi_{\text{old}}(y_{i,t} | x, y_{i,<t};\mathcal{R})},
\end{equation}
where $\hat{A}_{i,t}$ is the standardized advantage computed from group-relative rewards, $\epsilon$ controls the trust region size, and $\beta$ weights the KL penalty term. 


\heading{Multi-objective optimization}
After training the two models $\theta_{\text{forward}}$ and $\theta_{\text{backward}}$, we seek a balanced solution that integrates the strengths of both reward directions.
Inspired by linear mode connectivity \cite{neyshabur2020being,frankle2020linear}, we apply linear weight interpolation to combine the parameters of the two models, enabling the resulting model to simultaneously incorporate forward and backward reasoning capabilities. The final interpolated model $\theta_{\text{Bi-RAR}}$ is defined as: 
\begin{equation}
\theta_{\text{Bi-RAR}} = (1-\lambda) \cdot \theta_{\text{forward}} + \lambda \cdot \theta_{\text{backward}}, \quad \lambda \in [0,1],
\end{equation} 
where $\lambda$ controls the interpolation ratio.
By varying $\lambda$, we can explore a continuum of models that trade off between answer accuracy and question relevance, allowing flexible adaptation to different task requirements without the need for additional retraining.





\begin{table*}[t]
\centering
\renewcommand{\arraystretch}{0.9}
\begin{tabular}{l c c c c c c c c}
\toprule
 & \multicolumn{3}{c}{\textbf{General QA}} & \multicolumn{5}{c}{\textbf{Multi-Hop QA}} \\ 
\cmidrule(r){2-4} \cmidrule(r){5-9} 
\textbf{Methods} & NQ & TriviaQA & PopQA & HotpotQA & 2Wiki & Musique & Bamboogle & Avg. \\
\midrule
\multicolumn{9}{l}{\emph{Reasoning without retrieval}}  \\
Direct Inference & 0.106 & 0.288 & 0.108 & 0.149 & 0.244 & 0.020 & 0.024 & 0.134 \\
CoT  & 0.023 & 0.032 & 0.005 & 0.021 & 0.021 & 0.002 & 0.000 & 0.015 \\
SFT & 0.249 & 0.292 & 0.104 & 0.186 & 0.248 & 0.044 & 0.112 & 0.176 \\
R1-base  & 0.226 & 0.455 & 0.173 & 0.201 & 0.268 & 0.055 & 0.224 & 0.229 \\
R1-instruct  & 0.210 & 0.449 & 0.171 & 0.208 & 0.275 & 0.060 & 0.192 & 0.224 \\
\midrule
\multicolumn{9}{l}{\emph{One-step reasoning with retrieval}} \\
RAG  & 0.348 & 0.544 & 0.387 & 0.255 & 0.226 & 0.047 & 0.080 & 0.270 \\
\midrule
\multicolumn{9}{l}{\emph{Multi-step reasoning with retrieval}} \\
IRCoT  & 0.111 & 0.312 & 0.200 & 0.164 & 0.171 & 0.067 & 0.240 & 0.181 \\
Search-o1  & 0.238 & 0.472 & 0.262 & 0.221 & 0.218 & 0.054 & 0.320 & 0.255 \\
Search-R1-base & 0.421 & 0.583 & 0.413 & 0.297 & 0.274 & 0.066 & 0.128 & 0.312 \\
Search-R1-instruct & 0.397 & 0.565 & 0.391 & 0.331 & 0.310 & 0.124 & 0.232 & 0.336 \\
\hdashline\noalign{\vskip 0.6mm}
Bi-RAR-base & \textbf{0.442} & \textbf{0.614} & \textbf{0.432} & 0.317 & 0.297 & 0.073 & 0.188 & 0.338 \\
Bi-RAR-instruct &  0.438 & 0.608 & 0.421 & \textbf{0.391} & \textbf{0.402} & \textbf{0.153} & \textbf{0.363} & \textbf{0.397} \\
\bottomrule
\end{tabular}
\caption{Main results of Bi-RAR and baselines on QA benchmarks. The best performance is highlighted in bold.}
\label{tab:main_results}
\end{table*}

\section{Experimental Settings}

\heading{Datasets} We evaluate Bi-RAR on seven question answering benchmarks split into two groups: 
\begin{enumerate*}[label=(\roman*)] 
\item \textbf{General QA} datasets focus on factual questions that require accurate retrieval and understanding of real-world knowledge, generally involve single-hop reasoning:
NQ \cite{kwiatkowski2019natural}, TriviaQA \cite{joshi2017triviaqa}, and PopQA \cite{mallen2022not}.
\item \textbf{Multi-hop QA} datasets are specifically designed to evaluate a model’s ability to integrate multiple pieces of evidence across documents to answer a question, making them ideal for testing complex reasoning: 
HotpotQA \cite{yang2018hotpotqa}, 2WikiMultiHopQA \cite{ho2020constructing}, Musique \cite{trivedi2022musique}, and Bamboogle \cite{press2022measuring}.
\end{enumerate*}


\heading{Baselines} The baselines are grouped by how they incorporate retrieval into the reasoning process:
\begin{enumerate*}[label=(\roman*)] 
\item \textbf{Reasoning without retrieval}: These methods rely solely on the model's parametric knowledge to perform reasoning without retrieval, including Direct inference, Chain-of-Thought (CoT) reasoning \cite{wei2022chain}, Supervised fine-tuning (SFT) \cite{chung2024scaling} and RL-based fine-tuning without retrieval (R1) \cite{guo2025deepseek}.
\item \textbf{One-step retrieval and reasoning}: These approaches retrieve external evidence once before generating the answer, including Retrieval-Augmented Generation (RAG) \cite{lewis2020retrieval}.
\item \textbf{Multi-step retrieval and reasoning}: These methods perform iterative retrieval interleaved with reasoning, enabling the model to gather new information at each step, including IRCoT \cite{trivedi2023ircot}, Search-o1 \cite{li2025searcho1}, and Search-R1 \cite{jin2025searchr1} trained with GRPO.  
\end{enumerate*}
All baseline results are taken from Search-R1.
To ensure a fair comparison, all methods use the same retriever, retrieval setting, knowledge corpus, training dataset, and pre-trained LLMs.

\heading{Model variants} Bi-RAR includes two variants: 
\begin{enumerate*}[label=(\roman*)] 
\item \textbf{Forward-RAR}, trained only with the forward reward $R_{\text{forward}}$, as in $\theta_{\text{forward}}$; and 
\item \textbf{Backward-RAR}, trained only with the backward reward $R_{\text{backward}}$, as in $\theta_{\text{backward}}$. 
\end{enumerate*}
For these two variants, we only train a single model using the corresponding rewards, without performing interpolation.

\heading{Implementation details} We use both the Qwen2.5-3B-Base and Qwen2.5-3B-Instruct model \cite{qwen2025qwen25technicalreport}. 
Following Search-R1 \cite{jin2025searchr1}, we train our model on a combined training set of NQ and HotpotQA, adopt the same training and evaluation prompt template as Search-R1, and use \textbf{Exact Match (EM)} as the evaluation metric. 
For retrieval, we adopt the 2018 Wikipedia dump \cite{karpukhin2020dense} as the knowledge source and use E5 \cite{wang2022text} to simulate a search engine.

The training batch size is set to 128, and the validation batch size is set to 256, using only one-fourth of the training data compared to Search-R1.
To manage memory usage efficiently, we use gradient checkpointing and fully sharded data parallel (FSDP) with CPU offloading. 
For efficient response generation, we use vLLM with a tensor parallel size of 1 and a GPU memory utilization ratio of 0.6. 
Sampling is performed with a temperature of 1.0 and top-p of 1.0. 
We set the KL divergence regularization coefficient to $\beta = 0.001$, and the clipping ratio to $\epsilon = 0.2$.
In GRPO training, we follow the implementation from Verl \cite{sheng2025hybridflow}. 
Training runs for 200 steps. 
We set the policy model's learning rate to 1e-6 and sample 5 responses per prompt.
In multi-objective optimization, we tested $\lambda$ values of 0.25, 0.5, and 0.75, which emphasize different objectives.
We select $\lambda=0.25$ for both the Qwen2.5-3B-Base and Qwen2.5-3B-Instruct models as it achieved the best performance.

\begin{table*}[t]
\centering
\begin{tabular}{l c c c c c c c c}
\toprule
\textbf{Methods} & NQ & TriviaQA & PopQA & HotpotQA & 2Wiki & Musique & Bamboogle & Avg. \\
\midrule
\textbf{Qwen2.5-3B-Base} & & & & & & & & \\
Forward-RAR-base  & 0.440 & 0.613 & \textbf{0.435} & 0.313 & 0.293 & 0.066 & 0.169 & 0.333 \\
Backward-RAR-base & 0.435 & 0.599 & 0.423 & 0.313 & 0.282 & 0.069 & 0.125 & 0.321 \\
\hdashline\noalign{\vskip 0.6mm}
Bi-RAR-base & \textbf{0.442} & \textbf{0.614} & 0.432 & \textbf{0.317} & \textbf{0.297} & \textbf{0.073} & \textbf{0.188} & \textbf{0.338} \\
\midrule
\textbf{Qwen2.5-3B-Instruct} & & & & & & & & \\
Forward-RAR-instruct  & 0.432 & 0.598 & 0.418 & 0.376 & 0.375 & 0.144 & 0.347 & 0.384 \\
Backward-RAR-instruct & 0.436 & 0.602 & 0.391 & 0.380 & 0.339 & 0.145 & 0.347 & 0.377 \\
\hdashline\noalign{\vskip 0.6mm}
Bi-RAR-instruct &  \textbf{0.438} & \textbf{0.608} & \textbf{0.421} & \textbf{0.391} & \textbf{0.402} & \textbf{0.153} & \textbf{0.363} & \textbf{0.397} \\
\bottomrule
\end{tabular}
\caption{Ablation results of forward, backward, and bidirectional reasoning in Bi-RAR with different backbone LLMs.}
\label{table:Ablation study}
\end{table*}

\vspace{-2mm}
\section{Experimental Results}
In this section, we report the experimental results to demonstrate the effectiveness of Bi-RAR.

\subsection{Main results}
Table~\ref{tab:main_results} presents the overall performance of Bi-RAR compared to baseline methods across seven question answering benchmarks.
Observations on the baselines are:
\begin{enumerate*}[label=(\roman*)] 
\item Overall, models equipped with retrievers achieve better performance than those without, indicating that access to external knowledge sources can effectively complement the model's internal knowledge. 
\item Models perform better on general QA datasets than multi-hop QA datasets.
This discrepancy indicates that multi-hop reasoning and evidence aggregation remain challenging for the models.
\item Among the baselines, Search-R1 performs best, benefiting from iterative retrieval and outcome-based supervision that improve accuracy.
\end{enumerate*}

When we look at Bi-RAR, we find that:
\begin{enumerate*}[label=(\roman*)] 
\item Bi-RAR achieves the best overall performance among all evaluated models, with an average relative improvement of 18.2\% and 8.3\% over the strongest baseline when using Qwen2.5-3B Instruct and Base, respectively.
This demonstrates that our multi-objective optimization approach based on bidirectional information quantification supervision effectively constrains the reasoning trajectory, guiding the model to generate more accurate and compact answers.
\item Compared to the strongest baseline Search-R1, Bi-RAR delivers consistent gains across diverse datasets, despite being trained on only one-fourth of Search-R1's training data.
For example, Bi-RAR improves the performance on HotpotQA by 0.06 and 2Wiki by 0.092, corresponding to 18.1\% and 29.7\% relative increase, under the instruct-tuned model.
This indicates that bidirectional distances offer precise step-level optimization signals, leading to more efficient training and better inference quality.
\item Bi-RAR demonstrates effectiveness on both base and instruction-tuned models, suggesting strong generalization across model types.
\end{enumerate*}
\begin{figure*}[t]
    \centering
\vspace*{-2mm}
    \includegraphics[width=\linewidth]{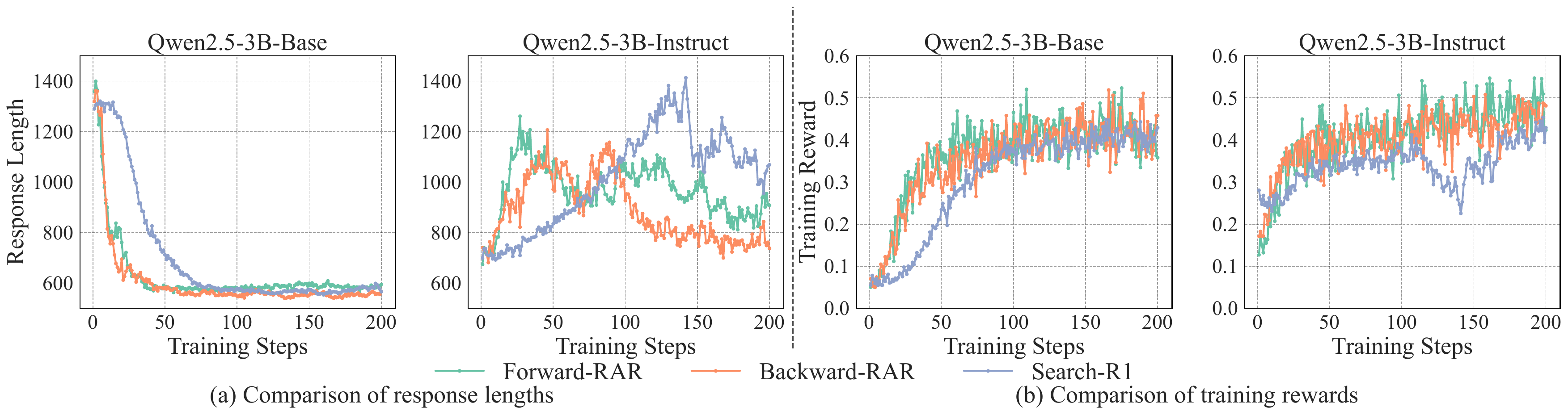}
    \vspace{-4mm}
    \caption{Trends in response lengths and rewards change during RL training for Forward/Backward-RAR and Search-R1.}
    \label{fig: length and reward}
\vspace*{-4mm}
\end{figure*}

\vspace{-2mm}
\subsection{Ablation study}
We conduct ablation experiments comparing the variants of Bi-RAR.
The results shown in Table~\ref{table:Ablation study} demonstrate that:
\begin{enumerate*}[label=(\roman*)] 
\item Forward-RAR performs better than Backward-RAR, with relative improvements of 3.7\% and 1.9\% on the base and instruct variants.
This result indicates that reward signals propagated from the answer side contribute more directly to final answer correctness. 
This aligns with our intuition, as anchoring generation on the expected answer better constrains the reasoning path.
\item Bi-RAR achieves the best performance across most datasets under both base and instruct backbone models.
This demonstrates the effectiveness of our multi-objective optimization framework in integrating forward and backward objectives, allowing the model to incorporate complementary reasoning signals and achieve stronger overall accuracy.
\end{enumerate*}

\vspace{-1mm}
\subsection{Training analysis}
We compare the training dynamics of Forward-RAR, Backward-RAR, and Search-R1 on both the Qwen2.5-3b-Base and Qwen2.5-3b-Instruct models, focusing on response length and train reward trends.

\heading{Response length} As shown in Figure~\ref{fig: length and reward}(a), on models initialized from the base model, the response lengths of Forward-RAR and Backward-RAR decrease faster than Search-R1. 
This demonstrates that the cascading reward structure, which emphasizes early trajectory alignment, leads to more efficient reasoning by guiding the model to eliminate unnecessary steps early in training.
On the instruction-tuned models, all methods show an initial increase followed by a decrease in response length. 
This is because the instruct model has a stronger instruction following ability, initially attempts to find correct answers via longer reasoning chains. 
As training progresses, the model learns that shorter responses can omit redundant steps while improving answer accuracy, leading to a response length reduction.
By the end of training, both Forward-RAR and Backward-RAR produce shorter responses than Search-R1 on the instruct model, indicating that our forward and backward information distance supervision effectively guides the model to generate accurate and concise reasoning steps.

\heading{Rewards} As shown in Figure~\ref{fig: length and reward}(b),
Forward-RAR and Backward-RAR converge faster than Search-R1 on both Base and Instruct models. 
This suggests that forward and backward reward signals provide more precise guidance for optimizing each step, enabling more effective training supervision.
On instruction-tuned models, the rewards of  Forward/Backward-RAR exhibit no severe fluctuations as observed in Search-R1 during training, which reflects the robustness and consistency of our bidirectional reward design.


\begin{figure}[t]
    \centering
    \includegraphics[width=\linewidth]{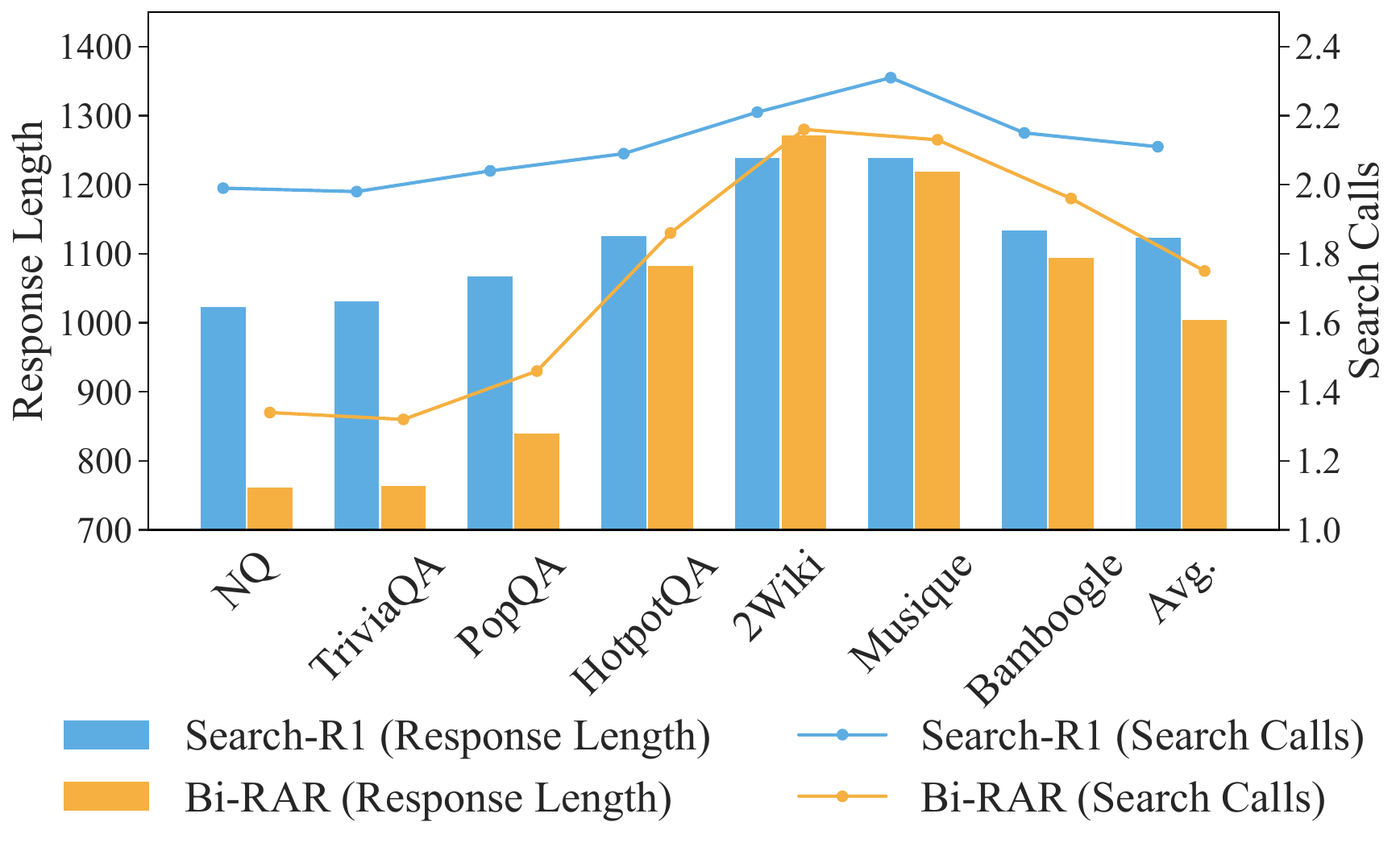}
    \vspace{-4mm}
    \caption{Response lengths and search calls in inference.}
    \label{fig: Evaluation-Analysis}  
    \vspace{-4mm}
\end{figure}

\vspace{-1mm}
\subsection{Inference analysis}
\label{sec:Efficiency Analysis}
To analyze the response efficiency during the inference phase, we compare Bi-RAR and Search-R1 in terms of response length and number of search calls, both of which directly affect inference efficiency.
We used the Qwen2.5-3B-Instruct model for comparison, with similar observations on other backbones.
The results across seven datasets and their average are shown in Figure~\ref{fig: Evaluation-Analysis}.

We observe that:
\begin{enumerate*}[label=(\roman*)] 
\item For response length, Bi-RAR generates shorter responses than Search-R1 on most datasets, with the reduction notable on the general QA datasets.
This is attributed to the cascading reward structure that emphasizes early trajectory alignment, enabling Bi-RAR to generate more concise and less redundant responses.
\item For search calls, Bi-RAR reduces the number of retrievals compared to Search-R1 across all datasets. 
This is because the bidirectional distances supervision mitigates invalid reasoning paths and corresponding redundant searches, leading to more efficient inference.
\item On 2WikiMultiHopQA, Bi-RAR produces longer responses than Search-R1 while using fewer search calls. 
This is due to the complex multi-hop reasoning required by the dataset, where our bidirectional supervision better guides the model to maintain coherent long-range inference with fewer but more targeted retrievals. 
As a result, Bi-RAR achieves a substantial relative performance gain of 29.7\%.
\end{enumerate*}





\vspace{-1mm}
\section{Related Work}
\heading{Retrieval-augmented generation} 
Retrieval-augmented generation \cite{lewis2020retrieval} is a widely adopted framework that enhances large language models (LLMs) \cite{achiam2023gpt,team2024gemini,LLMSurvey} by incorporating external knowledge sources. 
This technique effectively reduces hallucination \cite{zhang2023siren,liu2024robust,liu2025attack,liu2025scaling} and improves task performance \cite{gao2023retrieval,shuster2021retrieval,jiang2023activeretrievalaugmentedgeneration,shi2025search,li2025towards}. 
Building on this foundation, many studies have explored improving the performance of RAG systems by optimising prompts or training objectives, such as Self-RAG, REPLUG, and RA-DIT \cite{asai2023Self-RAG, shi2023REPLUG, lin2024RA-DIT}.
However, this single-round framework of retrieval-then-answering makes LLMs difficult to capture sufficient information and perform complete reasoning, leading to poor performance in handling complex multi-hop reasoning tasks.
To address this, recent approaches incorporate multi-step reasoning and retrieval, retrieval-augmented reasoning, to further enhance the model’s capability in complex scenarios:
\begin{enumerate*}[label=(\roman*)] 
\item IRCoT \cite{trivedi2023ircot} interleaves retrieval within the chain-of-thought reasoning process;
\item Search-o1 \cite{li2025searcho1} enhances LLMs by integrating agentic search capabilities that dynamically retrieve and incorporate external knowledge during the reasoning process;
\item Search-R1 \cite{jin2025searchr1} uses reinforcement learning to train LLMs to autonomously generate search queries and use real-time retrieval during step-by-step reasoning.
\end{enumerate*} 

These methods are primarily guided by the final answer, encouraging LLMs to interact more with search engines.
However, such a ``distant'' supervision signal cannot provide precise guidance for each interaction, leading to over or distorted reasoning directions by LLMs.
An effective reasoning trajectory should continuously progress toward the solution while remaining grounded in the original problem context.

Therefore, we propose a bidirectional information quantification to define the optimization objective for each reasoning step, enabling LLMs to determine the reasoning direction based on the current information sufficiency.

\heading{LLMs and reinforcement learning} Reinforcement learning \cite{kaelbling1996reinforcement} has fundamentally reshaped how we align LLMs with human preferences, evolving from computationally intensive strategies to more elegant and efficient solutions.
Early implementations such as PPO required both a reward and a critic model \cite{ouyang2022training,schulman2017proximal}.
DPO simplified this process by removing the reward model and directly optimizing on preference data \cite{rafailov2023direct}, while GRPO further simplifies the pipeline by dropping the critic model and using sampled responses to estimate advantages \cite{shao2024deepseekmath}.
These advances have significantly enhanced LLM reasoning capabilities, as evidenced by models such as OpenAI's o1, DeepSeek-R1 and Qwen2.5 \citep{jaech2024openai,guo2025deepseek,qwen2025qwen25technicalreport}.

In practice, LLM training often involves multiple optimization objectives that need to be effectively balanced during reinforcement learning.
Multi-objective reinforcement learning (MORL) \cite{barrett2008learning,roijers2013survey,li2020deep} extends standard RL by replacing the single scalar reward signal with multiple feedback signals, each corresponding to a different objective. 
Recent work \cite{rame2023rewarded} has begun exploring multi-objective optimization for LLMs. 

In this paper, we adopt a multi-objective reinforcement learning approach to simultaneously optimize the forward and backward objectives in our framework to achieve an effective balanced solution.




\vspace{-2mm}
\section{Conclusion}
We have proposed Bi-RAR, a novel retrieval-augmented reasoning framework designed to enhance multi-step reasoning ability of LLMs.
We introduce bidirectional information quantification grounded in Kolmogorov complexity theory, which jointly measures how far each reasoning step is from the final answer and how well it addresses the original question.
To effectively use these signals, we adopt a multi-objective reinforcement learning framework, enabling a smooth trade-off between the two objectives.
Experiments demonstrate that bidirectional reasoning guidance can significantly improve the accuracy of LLMs in solving complex problems, while achieving efficient interaction and reasoning with the search engine during training and inference.


\heading{Broader impact and limitations} 
We aim to make an initial exploration into multi-step retrieval-augmented reasoning, and to inspire the community to further advance this line of research.
As to the limitations of our work, we approximate Kolmogorov complexity using generation probabilities from an LLM when computing bidirectional information quantification, which could be time-consuming.
In future work, we plan to explore more efficient approximation methods and extend our framework to larger models.
Investigating efficient reasoning through low-resource model training in complex real-world search scenarios represents another promising direction.

\section*{Acknowledgments}
This work was funded by the Strategic Priority Research Program of the CAS under Grants No. XDB0680102, the National Natural Science Foundation of China (NSFC) under Grants No. 62472408 and 62441229, the National Key Research and Development Program of China under Grants No. 2023YFA1011602, the Dutch Research Council (NWO), under project numbers 024.004.022, NWA.1389.20.183, and KICH3.LTP.20.006, and the European Union under grant agreements No. 101070212 (FINDHR) and No. 101201510 (UNITE). All content represents the opinion of the authors, which is not necessarily shared or endorsed by their respective employers and/or sponsors.

\bibliography{aaai2026}

\end{document}